\pdfoutput=1

\documentclass[runningheads]{llncs}
\usepackage{graphicx}
\usepackage{tabularx}
\usepackage[ruled,vlined,linesnumbered]{algorithm2e}
\usepackage{subcaption}
\usepackage{multirow}
\usepackage{todonotes}
\usepackage[super]{nth}
\usepackage[backend=biber, style=nature]{biblatex}
\begin{document}

\SetKwComment{Comment}{\# }{}

\title{Streaming detection of significant delay changes in public transport systems}
\author{Przemysław Wrona\inst{1}\orcidID{0000-0002-5479-4489}\and
Maciej Grzenda\inst{1}\orcidID{0000-0002-5440-4954} \and
Marcin Luckner\inst{1}\orcidID{0000-0001-7015-2956}}
\institute{Warsaw University of Technology, Faculty of Mathematics and Information Science,\\
ul. Koszykowa 75, 00-662 Warszawa, Poland\\
\email{\{P.Wrona,M.Grzenda,M.Luckner\}@mini.pw.edu.pl}}

\maketitle              %
\begin{abstract}
Public transport systems are expected to reduce pollution and contribute to sustainable development. However, disruptions in public transport such as delays may negatively affect mobility choices. 
To quantify delays, aggregated data from vehicle locations systems are frequently used. However, delays observed at individual stops are  caused inter alia by fluctuations in running times and propagation of delays occurring in other locations. Hence, in this work, we propose both the method detecting significant delays and reference architecture, relying on stream processing engines, in which the method is implemented. The method can complement the calculation of delays defined as deviation from schedules. This provides both online rather than batch identification of significant and repetitive delays, and resilience to the limited quality of location data. 
The method we propose can be used with different change detectors, such as ADWIN, applied to location data stream shuffled to individual edges of a transport graph. It can detect in an online manner at which edges statistically significant delays are observed and at which edges delays arise and are reduced. Detections can be used to model mobility choices and quantify the impact of repetitive rather than random disruptions on feasible trips with multimodal trip modelling engines. The evaluation performed with the public transport data of over 2000 vehicles confirms the merits of the method and reveals that a limited-size subgraph of a transport system graph causes statistically significant delays
\end{abstract}
\section{Introduction}
Public transport (PT) is expected to contribute to sustainable development by reducing pollution and road congestion. However,  disruptions may negatively affect nominal and perceived journal time~\cite{YAP2021}. Hence, disruptions such as delays or lost transfers have been quantified to measure the performance of a PT system. Importantly, frequent public transport disruptions may negatively affect mobility choices. 
Developments in automatic vehicle location (AVL) \cite{Szymanski2018} systems
have largely increased the availability of spatio-temporal datasets documenting both the location of individual vehicles and real arrival and departure times. 
Such data is typically used for real time monitoring of public transport services, and improved operations~\cite{Raghothama2016}. 
Public transport schedules, such as schedules published in General Transit Feed Specification (GTFS) format can be compared against real departure times of PT vehicles. This provides for the aggregation of delays. As an example, in~\cite{Raghothama2016} 
delays at individual stoppoints were aggregated to provide features such as total delays per a stoppoint, the number of times a bus was delayed at a stoppoint and the average delay. Next, histograms of delays and maps of locations with delays and significant delays were produced. Recently, the newly available large volumes of delay report records were used for more in-depth analysis of delay data. In \cite{Szymanski2018}, a proposal to discretise delay changes into hour time bins and delay time bins was made. This is to consider and normalise the values associated with each bin separately. Importantly the calculations were performed for each edge representing a sequence of two stops consecutively visited by a vehicle. 
Agglomerative clustering was used to identify clusters of  edges of the PT system grouping edges similar in terms of delays observed at these edges.

Importantly, the majority of the works on the long term delay analysis rely on batch processing of data sets collected in the preceding periods.
In \cite{Raghothama2016}, this included hierarchical clustering and non-linear regression for delay prediction. In~\cite{Szymanski2018}, agglomerative clustering of edges was used, which provided for the identification of stop pairs between which minor or major delay changes  were observed under different probabilities throughout the entire period under consideration.

Many studies on long term analysis were based on averaging delay data.
However,  delays in a PT system may occur due to various reasons such as traffic light conditions preventing a vehicle from passing the crossroads, accidents, road reconstruction, too demanding schedules or demand fluctuations affecting boarding times at individual stops. Furthermore, some of the delays may be reported due to limited precision of location data obtained from GPS receivers and wrongly suggesting that a vehicle has not yet (or has already) departed from the stop. Hence we propose a Streaming Delay Change Detection (SDCD) method. The method can be used with varied change detectors applied to delay data to identify how frequently statistically significant delays occur at individual edges of a PT. The SDCD method we propose relies on stream processing, i.e. identifying changes in delay distribution in near-real-time rather than through batch processing of historical data and can be used with high volume data streams.

The primary contributions of this work are as follows:
\begin{itemize}
    \item We propose the SDCD method to monitor and detect changes in delay distribution, 
    and propose two variants of the method to detect changes during entire days and individual time slots.
    \item We evaluate the method with real Warsaw public transport data and make the implementation of  the method and data available for the research community\footnote{The source code of the SDCD method and other resources related to this work are available at \url{https://github.com/przemekwrona/comobility-sdcd}}.
\end{itemize}

\section{Related works}
\subsection{Quantifying delays and change detectors}
Delays in public transport systems are typically analysed based on the data sets aggregating delays observed for individual vehicles at stop points such as bus stops~\cite{Luckner2020a,Raghothama2016} or edges defined by two consecutive stop points~\cite{Szymanski2018}. Some studies go beyond calculating average delay values.
As an example, Szymanski et al. proposed using bins of variable lengths for aggregating delay values e.g. grouping delays of [-10.5 min,-5.5 min] in a single bin~\cite{Szymanski2018}. 

 Yap et al. in \cite{Yap2021} note the difference between the change in PT system performance caused by stochastic demand or supply fluctuations i.e. the change referred to as  disturbance, and  disruption, which is the change caused by distinctive incidents or events. Both these changes are examples of perturbations. Importantly, disruptions can propagate in the PT system  and their consequences can be observed even in distant locations. Due to complex demand-supply interactions, in the case of urban PT networks, simulations-based models are often necessary to predict the impact of disruptions \cite{Yap2021}.

The volume of  vehicle location data collected from AVL systems is growing. It reached 12 mln records reported in a study for Stockholm~\cite{Raghothama2016}, 16 mln for Wroclaw \cite{Szymanski2018} or even 2.9 bln of records  collected for Warsaw over approximately 30 months~\cite{Luckner2020a}.
This
inspired research into the use of big data frameworks for the storage and processing of location and delay records. A survey of related works and a proposal for a unified architecture serving storage and analytical needs of IoT data with emphasis on vehicle location data can be found in~\cite{Luckner2020a}.

In parallel, developments in stream rather than batch processing of high volume and velocity data raised interest in change detection methods applicable to data streams.
One of the popular detectors is ADWIN proposed in 
\cite{Bifet2007}.
In ADWIN, the adaptive window approach is used for streaming data and applied to detect changes in the average value of a stream of bits or real-valued numbers.
The size of sliding windows used for change detection is not constant and defined a priori, but depends on the rate of detections. Thus for stationary data, the window is growing, but in the case of detection, the window is narrowing to discard historical data. The only parameter of the detector is a confidence value $\delta\in(0,1)$, controlling the sensitivity of the detection, i.e. influencing the ratio of false positives. A change is detected with a probability of at least $1-\delta$ .

Another recently proposed approach to concept drift detection relies on the Kolmogorov-Smirnov test applied to sliding windows populated with recent data instances from a data stream~\cite{Raab2020}. The parameters for the {\bf KSWIN} detector are the probability $\alpha$ of the test statistic and the sizes of two sliding windows used for the detection of a difference between the distributions of data present in the two windows. Concept change is detected when the distance between the empirical cumulative distribution functions (eCDFs) of the two differently sized windows exceeds the $\alpha$-dependant threshold. 

Research into change detectors is largely inspired by the need to detect when an update of the learning model is needed to adapt the model to concept drift. A family of methods monitoring the mean estimated from real values with an explicit focus on monitoring the values of performance measures of learning models was proposed in \cite{Blanco2015}. The methods rely on {\bf HDDM} algorithms proposed in the study and use Hoeffding's inequality to report warnings and actual drifts based on two confidence levels -- the parameters of the change detector. 

The change detection methods applicable to data streams were not used until very recently for transport data. Among the first works of this kind, Moso et al. in~\cite{Moso2020} addressed the problem of collecting message exchanges between vehicles and analysing trajectories. Variations of trajectories from normal ones were detected to identify anomalies. This recent study is among the first studies exploiting the use of Page-Hinkley and ADWIN change detection methods to process Cooperative Awareness Messages produced by vehicles in order to perform  road obstacle detection. Out of the two methods, ADWIN yielded promising results, which is unlike Page-Hinkley, which additionally required parameter tuning. %

\subsection{Analysing multi-modal connections and  the impact of perturbations on travel times}
Delays of individual PT vehicles not only have an impact on travel time, but also may cause lost transfers. As some trips require multiple connections and multi-modal routes, to estimate travel times under static schedules and real conditions, simulation software is needed. 
A popular solution is to use OpenTripPlanner (OTP)\footnote{\url{http://www.opentripplanner.org/}} - an open-source and cross-platform multi-modal route planner. OTP gives the ability to analyse varied transport data. That includes modifications of schedules (also in real-time) and changes to the street network. Importantly, OTP can model the effects of road infrastructure changes and examine the consequences of temporary changes in schedules~\cite{Young2021}.

Several recent scientific works used OTP as an analytic tool.
Lawson et al. examined a “blended data” approach, using an open-source web platform based on OTP to assist transit agencies in forecasting bus rider-ship~\cite{Lawson2021}. Ryan et al. used OTP to examine the critical differences between the two representations of accessibility, calculating door-to-door travel times to supermarkets and healthcare centres~\cite{Ryan2021}. To perform connection planning both static PT schedules made available in GTFS format~\cite{Waldeck2020,Ryan2021} and real feed of vehicle arrival and departure times in the form of GTFS Realtime\footnote{\url{https://developers.google.com/transit/gtfs-realtime}}~\cite{Liebig2014} can be used. In particular, a comparison of travel times estimated by OTP under planned schedules and real departure times provided in GTFS Realtime can be made.

While the stream of real arrival and departure times, including possible delays, can be forwarded to a modelling environment such as OTP, this does not answer whether delays exemplify systematic problems at some edges of the PT graph or occasional fluctuations.
Hence, in our study, we focus on detecting statistically significant perturbations in the performance of a public transport system. In this way, we aim to reduce the risk of reporting disruptions caused by stochastic fluctuations, unless these disruptions occur frequently.
Hence, rather than averaging delays possibly observed occasionally and caused by limited precision of GPS readouts, variability in the number of passengers or traffic light conditions, we aim to identify these delays which occur frequently and over longer periods. To make this possible, we propose a method applying change detectors to data collected at individual edges of public transport graph and the architecture within which the method can be implemented.

Furthermore, let us note that such detections can provide basis for generating schedules reflecting regular statistically significant delays and using them in a simulation environment such as OTP.  

\section{Architecture of delay detection and modelling system}
To validate the approach proposed in this work, we implemented SDCD method as a part of IoT platform collecting and analysing sensor data, including data from AVL systems. The platform we used for the collection and processing of vehicle location and delay data is an update and extension of  
USE4IoT architecture~\cite{Luckner2020a}. 
Let us note that without the loss of generality, by delays we mean both arrivals before and after scheduled time.
The USE4IoT is an Urban IoT platform designed as an extension of Lambda architecture. It fulfils the requirements of the Lambda pattern and adds extra layers matching the needs of smart cities.  

\begin{figure}[t]
\centering
\includegraphics[width=0.8\textwidth]{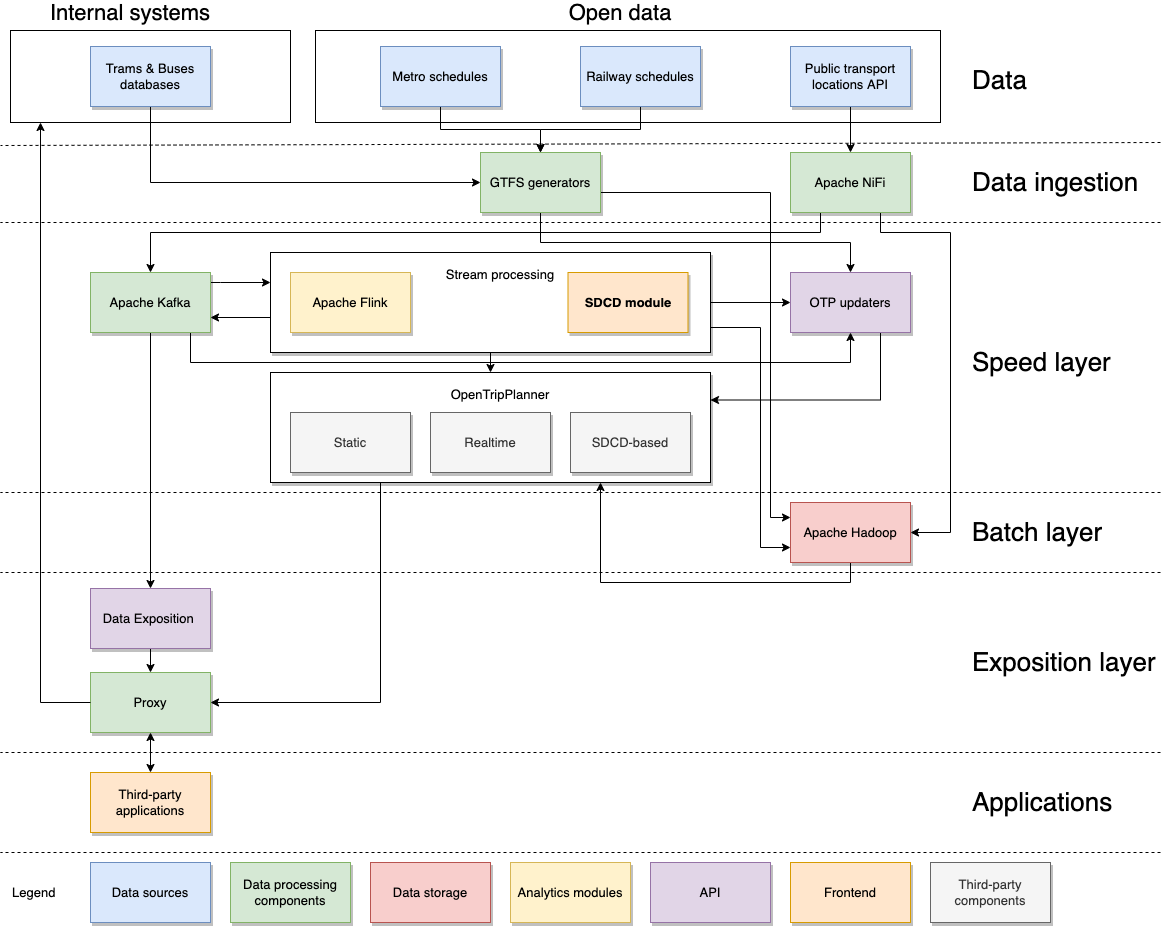}
    \caption{The architecture of delay detection and impact modelling system}
    \label{fig:Architecture}
\end{figure}

Fig.~\ref{fig:Architecture} presents the architecture of the part of the system related to the SDCD method. In the analysed case, input data comes from the open data portal of the city of Warsaw\footnote{\url{https://api.um.warszawa.pl}} and additional open sources.
Some data is collected online from a public transport localisation stream. Other data, such as timetables, are downloaded periodically. 
The data ingestion layer is responsible for collecting the data from the various data sources. It requires a combination of components and frameworks. In the case of USE4IoT, Apache NiFi is used to poll the data sources for the most recent data and convert new data records into data streams. 

Big data, including location data streams, are archived using the Hadoop Distributed File System to store tabular data, including data collected as online data streams and timetables downloaded daily. The vehicle location streams are redirected to stream processing engines through  Apache Kafka to ensure high throughput and resilience to downstream performance. Next, the Apache Flink application is used to process and merge the location of vehicles with PT timetables. The architecture provides stream processing with a mean delay of less than 2.5 seconds~\cite{Luckner2020a}. 

In this work, we propose three OTP instances, each serving different needs. The OTP instances are updated from three types of sources. Static GTFS data is created based on static schedules and uploaded into the first OTP instance. The stream analytics modules detect delays and untypical events and supplies OTP with a real-time GTFS. Therefore, the real-time OTP instance can calculate multi-modal connections considering current vehicle location and delays. Finally, we propose the SDCD module to detect statistically important changes in public transport delays. This can provide the basis for GTFS files containing credible schedules, i.e. the schedules reflecting statistically significant delays that update the static departure times, to be used in the SDCD-based OTP instance. In this way, a comparison between travel options and times under a) static schedules, b) real-time situation and c) schedules reflecting significant delays possibly observed frequently over preceding days can be made. In this way, the impact of delays on travel times and related aspects such as lost transfers between individual connections can be investigated.

Finally, the entire architecture was created to forward the results through the data exposition layer to the application layer. The processing results can be consumed by any application including applications created by third-party developers. However, the core part of the solution, which we focus on in this work, is the SDCD method providing the basis for online detection of statistically significant delay changes.

\section{Streaming delay change detection}

 Let $L$ denote the set of PT lines, each defined by a sequence of bus or tram stops $\{s_{l,1},\ldots,s_{l,j}\}$. Let us note that when describing PT system, we will rely on the notation similar to the notation proposed in \cite{Yap2021}. In our case, we assume that $s_{l,j}=s_{l,1}$ i.e. a line is defined by a loop, while stops visited in one direction are not necessarily the stops visited by a vehicle travelling in the opposite direction.

Let PT network be a directed graph $G=(S,E)$, where $S$ is the set of all stops in the network e.g. in the urban PT network and $E$ is the set of edges. An edge $(s_i,s_j), i\neq j$ exists i.e. $(s_i,s_j)\in E$ if and only if at least one line $l$ exists such that the two consecutive stops of the line are $s_i$ and $s_j$. 

Furthermore, let $\mathcal{S}_1,\mathcal{S}_2,\ldots$ be the stream of location records received over time from an AVL system. Without the loss of generality, we assume each $\mathcal{S}_i$ contains both current geocoordinates of a vehicle course $v$, the line $l$ operated by the vehicle, and the identifiers of two most recently visited  stops $s_{i}, s_{i-1}$ by the vehicle. Real departure times and planned departure times as defined in static schedules are also available for both of these stops. These are denoted by $t_\mathrm{R}(s,v)$ and $t_\mathrm{S}(s,v)$, respectively. 
Hence, $d(\mathcal{S}_j)=d(s_i,v)=t_\mathrm{R}(s_i,v)-t_\mathrm{S}(s_{i},v)$ denotes delay i.e. the difference between real and planned departure time for a vehicle course $v$ observed at stop $s_i$ i.e. the most recently left stop during course~$v$.
Let us note that if raw data from AVL include no line identifiers, they can be retrieved from schedule data. Furthermore, if needed stop identifiers can be identified based on past vehicle coordinates and stop coordinates of the line served by the vehicle. The data of vehicles not in service are skipped.

\begin{algorithm}[t]
\DontPrintSemicolon

\SetKwData{Left}{left}
\SetKwData{This}{this}
\SetKwData{Up}{up}
\SetKwFunction{Union}{Union}
\SetKwFunction{getNextVehicleSnapshot}{getNextVehicleSnapshot}
\SetKwFunction{isOnTheStop}{isOnTheStop}
\SetKwFunction{getDetector}{getDetector}
\SetKwFunction{putDetector}{putDetector}
\SetKwFunction{getDelayAtStop}{getDelayAtStop}
\SetKwFunction{addDelay}{addDelay}
\SetKwFunction{detectedChange}{detectedChange}
\SetKwFunction{save}{save}
\SetKwFunction{GetDetectorId}{GetDetectorId}

\SetKwInOut{Input}{input}

\SetKwInOut{Output}{output}
\Input{$stream$: a stream of vehicle locations $\mathcal{S}_1,\mathcal{S}_2,\ldots$ 
\newline $GetDetectorId(V)$: a function returning the identifier of the change detector to be used
\newline $\Delta \in\{\mathrm{TRUE},\mathrm{FALSE}\}$ - the parameter defining whether to process $\Delta d()$ or $d()$ delays}
\Output{$detections$: A stream of  detected changes in delay stream}

\BlankLine
$ConceptChangeDetectors \gets \{\}$ \Comment*[r]{Initialize empty map of detectors}
\BlankLine
$i=1$;\\
\While{stream has next element}{
    \BlankLine
    $V \gets \mathcal{S}_i$\;
        \BlankLine
        $K \gets \GetDetectorId(V)$\;
        $D \gets ConceptChangeDetectors.\getDetector(K)$\;
        \BlankLine
        \If{D is NULL} {
        $D \gets new ConceptChangeDetector()$\;
        $ConceptChangeDetectors.\putDetector(K, D)$\;
        }
        \BlankLine
        \If{$\Delta$}{
        $delay \gets \Delta d(V)$\;
        }
        \Else{
        $delay \gets d(V)$\;
        }
        $D.\addDelay(delay)$\;
        \BlankLine
        \If {D.\detectedChange()}{
            $detections.\save(V,D.identifier)$\;
        }
    $i=i+1$;
}
\caption{Streaming delay change detection
algorithm}\label{algo_delay_detection}
\end{algorithm}
Our approach to detect changes in a stream of delays uses change detectors such as detectors relying on the ADWIN algorithm \cite{Bifet2007}. The SDCD method is defined in Alg.~\ref{algo_delay_detection}.
As an input stream, we use the location stream of PT vehicles $\mathcal{S}_1,\mathcal{S}_2,\ldots$ described above. 
Furthermore, the location stream can be shuffled into substreams linked to individual edges of PT graph or bins linked to a combination of an edge and an hour $h=0,\ldots,23$ of the day. In the first case, which we call edge-based, all vehicle location records describing vehicles visiting the sequence of two stops defining an edge will be gradually processed by one change detector. In the bin-based approach, all records related to an edge and time of the day defined by a one-hour time slot will be processed together. Hence, the intuition behind the edge-based approach is to identify delays and delay reductions as they appear over time. In this case, the detector is recognised by pair of stops. Thus, for each pair of stops, one detector that collects data all the time is created.

We propose bin-based approach to identify possible changes in delays at the same time of the day, e.g. between 8:00 and 8:59 over consecutive days and occurring at one edge of PT graph. In this case, the detector identification is extended by the hour that comes from the current vehicle timestamp. Hence, at most 24 detectors are created for each pair of stops visited in a row.
The two approaches of defining detector keys are formally defined in Alg.~\ref{algo_detector_id}.

Moreover, 
we propose to calculate delay change between stops, defined as $\Delta d(\mathcal{S}_j)=d(s_i,v)-d(s_{i-1},v)$. Let us note that $d(\mathcal{S}_j)>0$ may be accompanied by $\Delta d(\mathcal{S}_j)=0$ or even $\Delta d(\mathcal{S}_j)<0$. As an example, it is possible that a delayed vehicle ($d(\mathcal{S}_j)>0$) has reduced its delay when travelling between stops $s_{i-1}$ and $s_i$ i.e. $\Delta d(\mathcal{S}_j)<0$. Hence, the third parameter of Alg.~\ref{algo_delay_detection} is whether to detect changes in $d()$ or $\Delta d()$ streams of values.

\begin{algorithm}
  \DontPrintSemicolon
  \SetKwFunction{getCurrentStopId}{getCurrStopId}
  \SetKwFunction{getPreviousStopId}{getPrevStopId}
  \SetKwFunction{DetectorId}{DetectorId}
  \SetKwFunction{getDetectorIdForEdge}{getDetectorIdForEdge}
  \SetKwFunction{getDetectorIdForEdgeAndTime}{getDetectorIdForEdgeAndTime}
  \SetKwFunction{getHour}{getHour}
  \SetKwFunction{join}{join}
  \SetKwProg{Fn}{Function}{:}{}
  \Fn{\getDetectorIdForEdge{$V$}}{
        \KwRet $V.\getCurrentStopId().\join(V.\getPreviousStopId())$\;
  }
  
  \SetKwProg{Gn}{Function}{:}{}
  \Gn{\getDetectorIdForEdgeAndTime{$V$}}{
        \KwRet $V.\getCurrentStopId().\join(V.\getPreviousStopId()).\join(V.\getHour())$\;
  }
  \caption{The functions calculating detector identifiers.
  }\label{algo_detector_id}
\end{algorithm}

During the algorithm initialisation, we create an empty map of detectors (Line 1). Every time data for a new detector key, i.e. new edge or new bin, is encountered in the stream for the first time, we create a new change detector object  (Lines 7-9).
Next, we add the value of delay expressed in seconds to the detector and check if the detector detected a change in the stream. Detected change is saved together with detector key, i.e. edge identifier in the case of edge-based, and edge and time slot in the bin-based approach.

Finally, once significant perturbations in the performance of a public transport system defined by repetitive detections of delays are identified, we propose to develop SDCD-based schedules, i.e. the public transport schedules reflecting significant perturbations observed at individual edges. Next, by comparing the behaviour of a public transport system under static schedules and SDCD-based schedules, the impact of significant perturbations  can be assessed. For example, bus delays may cause missing a scheduled connection at a transfer stop and largely increase overall travel time.

\section{Results}
\subsection{Reference data \label{sec:data}}

The data used to validate the methods comes from Warsaw Public Transport public API that provides the current position of vehicles every 10 seconds, which yields 2.0-2.5 GB of  data each day. The average daily number of records over the period selected to illustrate the results of this study exceeds 4 million (839 thousand for trams and 3.17 million for buses). The ratio of records linked with static schedule reached 92\%. 
The remaining records represent inter alia vehicles not in service, for which GPS readouts are also  available at a data source.

The public transport vehicles travel an average of 14.6 thousand edges $E$ daily, defined by two following stops.
An average edge of the public transport network is visited by 54 vehicles (the median) per day. However, the actual range is from a single vehicle course per edge to over ten thousand on some city centre edges.
The median delay at an edge reaches 104 seconds, which is considered acceptable according to criteria adopted by the local public transport authority.

\subsection{Change detections}

In the first experiments, Alg.~\ref{algo_delay_detection} was used with three change detectors -- ADWIN \cite{Bifet2007}, KSWIN  \cite{Raab2020}, and  HDDM~\cite{Blanco2015} used to perform change detection. In the case of HDDM methods, HDDM\_A was selected for the experiments. It relies on a lower number of parameters than HDDM\_W, which additionally requires weight given to recent data to be set. Setting such a weight would require additional hyperparameter tuning. Hence, for the sake of simplicity, HDDM and HDDM\_A will be used interchangeably in the remainder of this study.
In all experiments, the implementation of detectors from {\texttt scikit-multiflow} library was used, and  
default settings of the ADWIN detector were applied. In the case of KSWIN and HDDM, the same confidence setting as for ADWIN was applied.

\begin{figure*}[t] %
\centering
\begin{subfigure}[b]{0.33\textwidth}
    \includegraphics[width=\columnwidth]{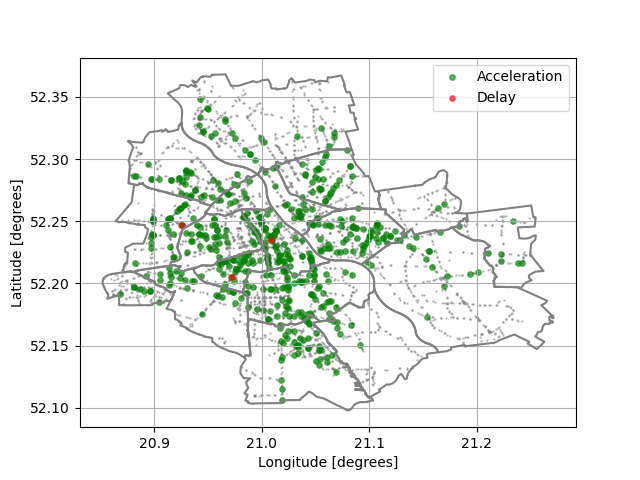}
    \caption{ADWIN $delay$}
    \label{fig:warsaw_adwin_delay_in_week}
\end{subfigure}
\begin{subfigure}[b]{0.32\textwidth}
    \includegraphics[width=\columnwidth]{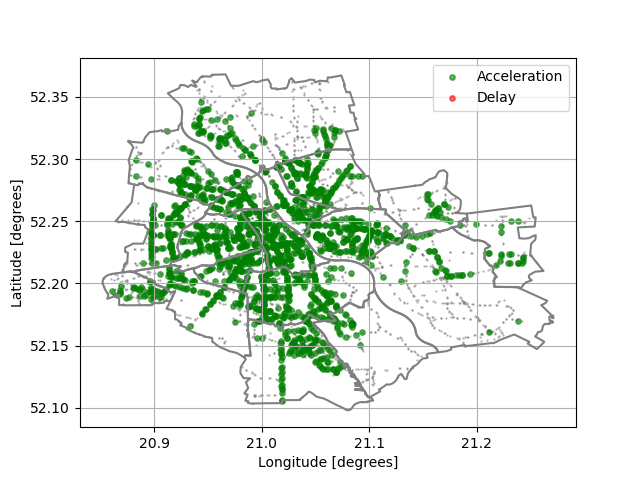}
    \caption{KSWIN $delay$}
    \label{fig:warsaw_kswin_delay_in_week}
\end{subfigure}
\begin{subfigure}[b]{0.33\textwidth}
    \includegraphics[width=\columnwidth]{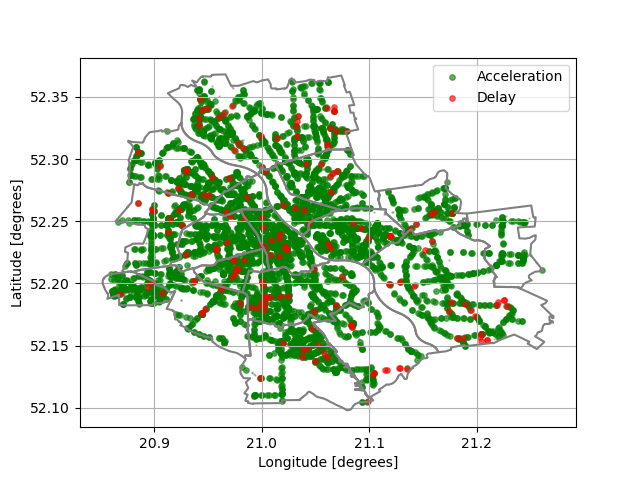}
    \caption{HDDM $delay$}
    \label{fig:warsaw_hddma_delay_in_week}
\end{subfigure}
\begin{subfigure}[b]{0.33\textwidth}
    \includegraphics[width=\columnwidth]{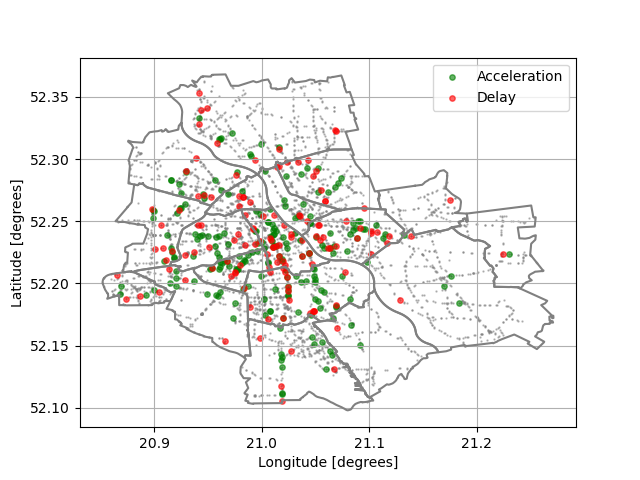}
    \caption{ADWIN $\Delta delay$}
    \label{fig:warsaw_adwin_delta_in_week}
\end{subfigure}
\begin{subfigure}[b]{0.33\textwidth}
    \includegraphics[width=\columnwidth]{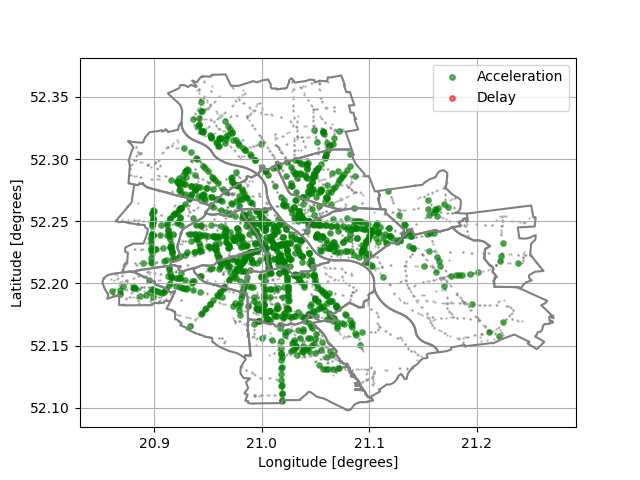}
    \caption{KSWIN $\Delta delay$}
    \label{fig:warsaw_kswin_delta_in_week}
\end{subfigure}
\begin{subfigure}[b]{0.32\textwidth}
    \includegraphics[width=\columnwidth]{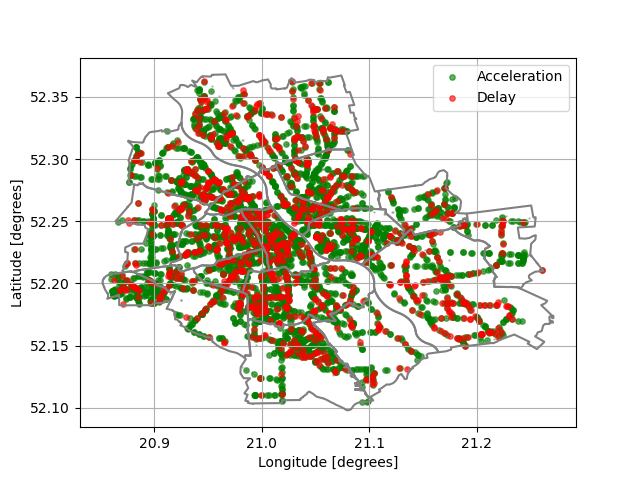}
    \caption{HDDM $\Delta delay$}
    \label{fig:warsaw_hddma_delta_in_week}
\end{subfigure}
\caption{The locations at which changes in delays were detected in the location stream between 1\nth{8} December 2021 and 2\nth{1} December 2021. Edge-based approach.}
\label{fig:warsaw_week}
\end{figure*}

Fig.~\ref{fig:warsaw_week} presents the edges detectors of which reported at least one delay change during the reference period selected to visualise the results of this study, i.e. 1\nth{8} December 2021 (Saturday) till 2\nth{1} December 2021 (Tuesday). 
The detectors are organised into two types. The first type of detectors analysed delay $d()$ at the destination stop $s_i$ of edge $e=(s_{i-1},s_{i})$, hereafter referred to as {\em delay}. The second type analysed a change of  delay $\Delta d()$ observed at an edge $e$, referred to as $\Delta${\em delay} in the remainder of this work. 
In this experiment, one detector analysed the data from entire days to find delay changes, i.e. edge-based approach was used. For the sake of clarity the edges at which detections occurred are depicted by points placed in the destination stop $s_i$ of edge $e=(s_{i-1},s_{i})$.

The ADWIN algorithm (Fig.~\ref{fig:warsaw_adwin_delay_in_week})
detected the smallest number of delay changes in comparison to KSWIN (Fig.~\ref{fig:warsaw_kswin_delay_in_week}) and HDDM
(Fig.~\ref{fig:warsaw_hddma_delay_in_week}). Interestingly, all algorithms detected more accelerations ($d()<0$) than delays ($d()>0$). It may look positive. Still, accelerations are rare compared to the number of all edges. A possible explanation of the larger number of  acceleration events than of delay events is that major delays are easy to attain at even short distances, but reducing them inevitably takes more time i.e. longer distances over which delay reduction has to be attained by the drivers, which is reflected by a larger number of accelaration edges.

The detection of delay changes $\Delta d()$ is more diverse. While the ADWIN  (Fig.~\ref{fig:warsaw_adwin_delta_in_week}) and HDDM
(Fig.~\ref{fig:warsaw_hddma_delta_in_week}) detect both directions of changes, the KSWIN (Fig.~\ref{fig:warsaw_kswin_delta_in_week}) detects mostly accelerations. It may be caused by the fact that the KSWIN is comparing eCDF functions while the other algorithms compare the mean values. Once again, the ADWIN algorithm detected the smallest number of changes.

\begin{figure*}[t] %
\centering
\begin{subfigure}[b]{0.33\textwidth}
    \includegraphics[width=\columnwidth]{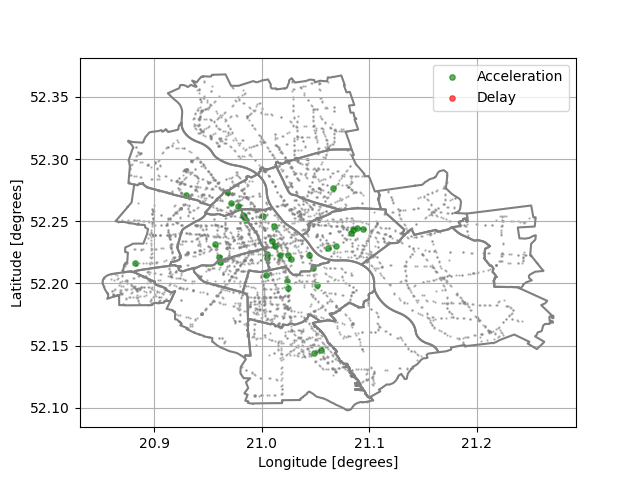}
    \caption{ADWIN $delay$}
    \label{fig:warsaw_adwin_delay_in_week_hourly}
\end{subfigure}
\begin{subfigure}[b]{0.33\textwidth}
    \includegraphics[width=\columnwidth]{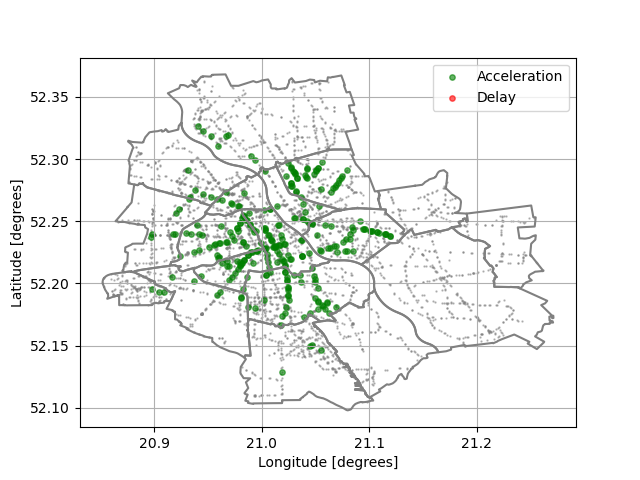}
    \caption{KSWIN $delay$}
    \label{fig:warsaw_kswin_delay_in_week_hourly}
\end{subfigure}
\begin{subfigure}[b]{0.32\textwidth}
    \includegraphics[width=\columnwidth]{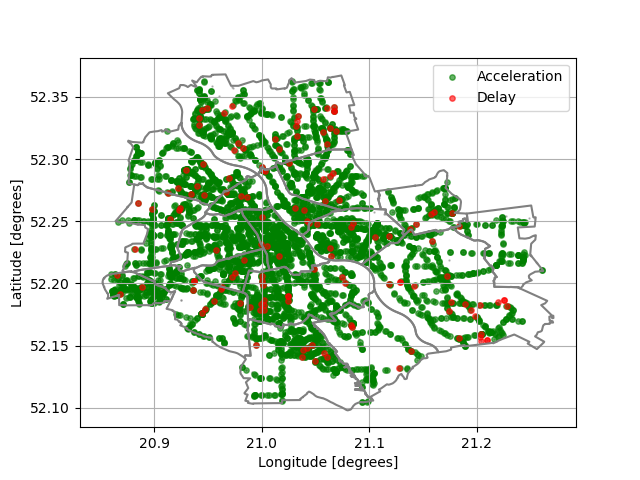}
    \caption{HDDM $delay$}
    \label{fig:warsaw_hddm_delay_in_week_hourly}
\end{subfigure}
\begin{subfigure}[b]{0.33\textwidth}
    \includegraphics[width=\columnwidth]{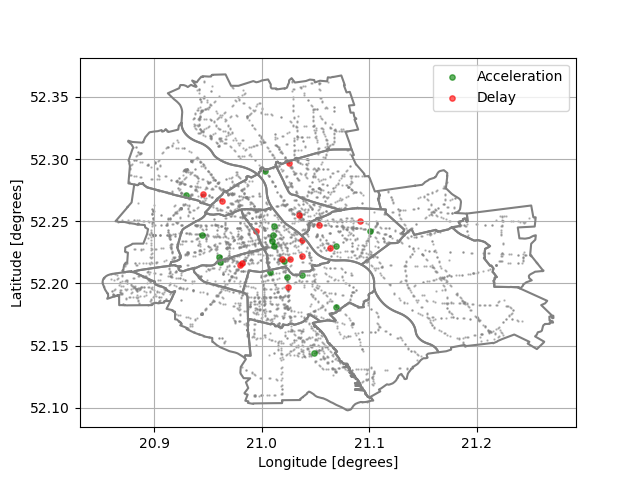}
    \caption{ADWIN $\Delta delay$}
    \label{fig:warsaw_adwin_delta_in_week_hourly}
\end{subfigure}
\begin{subfigure}[b]{0.33\textwidth}
    \includegraphics[width=\columnwidth]{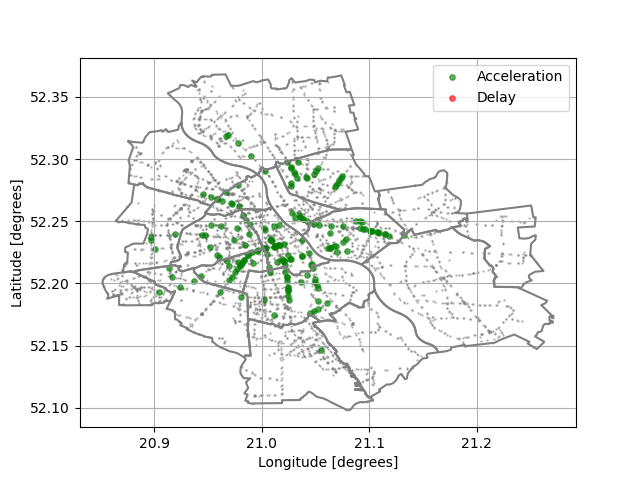}
    \caption{KSWIN $\Delta delay$}
    \label{fig:warsaw_kswin_delta_in_week_hourly}
\end{subfigure}
\begin{subfigure}[b]{0.32\textwidth}
    \includegraphics[width=\columnwidth]{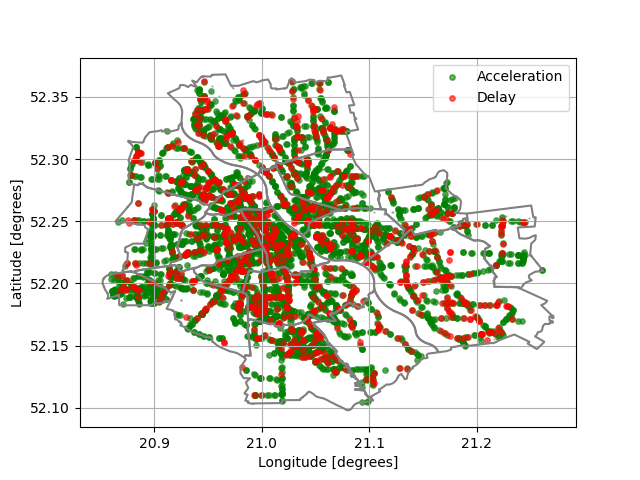}
    \caption{HDDM $\Delta delay$}
    \label{fig:warsaw_hddm_delta_in_week_hourly}
\end{subfigure}

\caption{The locations at which changes in delays were detected in the location stream between  1\nth{8} December 2021 and 2\nth{1} December 2021. Bin-based approach.}
\label{fig:warsaw_week_hourly}
\end{figure*}
In the second experiment -- instead of a single detector working throughout the entire period  --  the bin-based approach divided the records related to an edge into
one-hour time slots. The rest of the conditions stay the same as in the previous experiment. The results are presented in Figure~\ref{fig:warsaw_week_hourly}.
For the ADWIN (Fig.~\ref{fig:warsaw_adwin_delay_in_week_hourly} and Fig.~\ref{fig:warsaw_adwin_delta_in_week_hourly}) and KSWIN (Fig.~\ref{fig:warsaw_kswin_delay_in_week_hourly} and Fig.~\ref{fig:warsaw_kswin_delta_in_week_hourly}) algorithms the number of detected changes drops rapidly compared to edge-based approach. This effect shows that most detected delay changes are statistically important compared to other periods of the day but are rather typical for the specific hour, which sounds reasonable because of  dynamic traffic changes during the day. However, the drop out effect is not observed for the HDDM algorithm (Fig.~\ref{fig:warsaw_hddm_delay_in_week_hourly} and Fig.~\ref{fig:warsaw_hddm_delta_in_week_hourly}), which may suggests that the HDDM algorithm detects again too many events.

The ADWIN algorithm is the least demanding in the context of parametrisation. Moreover, the results of other algorithms are counterintuitive when KSWIN detects only accelerations or the bin-based approach does not reduce the number of HDDM detections. Therefore, the ADWIN was selected for further analysis.

Table~\ref{tab:summary} presents statistics for all ADWIN delay $d()$ and delay change $\Delta d()$ detections in the edge-based approach. The number of detections is relatively small compared to the daily throughput and the number of analysed edges. When one compares the medians, the detected delays $d()>0$ are comparable to the median delays of 104 seconds (see Section~\ref{sec:data}). Their standard deviation is relatively high and similar to the median. Therefore, the detected changes have a local character in the sense of a detection value (which would not necessarily be an exception in another location), but are globally shifted to reductions, which are taken as exceptions in contrast to the global delay level.

The proportion between reductions and increases is more balanced for $\Delta$delay change detections. %
A very small median and several times higher standard deviation reveals that many detections concern minor delay change only, %
which additionally helps focus on these edges at which major delay change occurs.

To sum up, the statistical results show that statistically significant delay changes are rare for thousands of analysed connections.
In practice, it is recommended to use both types of detections ($d()$ and $\Delta d()$) with an additional cut off of the small absolute values to focus on delays which are both statistically significant and high.

\begin{table}[t]
\centering
\caption{Delay changes detected with SDCD  algorithm.
ADWIN detector.}
\label{tab:summary}
\begin{tabular}{lrrrrrr}
\hline
\multicolumn{1}{c}{Delay type}&\multicolumn{1}{c}{Date}       & \multicolumn{1}{c}{Location records} & \multicolumn{1}{c}{Increases} &  \multicolumn{1}{c}{Reductions} & \multicolumn{1}{c}{Median[s]} & \multicolumn{1}{c}{STD[s]} \\ \hline
\multirow{4}{*}{$d()$}&2021-12-18 &       1181271 &               5 &             1059 &     131.0 &  125.0 \\ 
&2021-12-19 &       1242939 &              10 &              666 &     110.0 &   84.0 \\ 
&2021-12-20 &       1256871 &               7 &              862 &     107.0 &   91.0 \\ 
&2021-12-21 &       1049178 &               6 &              680 &     131.0 &  101.0 \\ \hline
\multirow{4}{*}{$\Delta d()$}&2021-12-18 &       1181271 &             249 &              365 &       6.0 &   32.0 \\ 
&2021-12-19 &       1242939 &             199 &              299 &       4.0 &   29.0 \\ 
&2021-12-20 &       1256871 &             219 &              336 &       5.0 &   55.0 \\ 
&2021-12-21 &       1049178 &             202 &              310 &       5.0 &   45.0 \\ \hline

\end{tabular}
\end{table}

\subsection{Peak hours analysis}

\begin{figure*}[t] %
\centering
\begin{subfigure}[b]{0.33\textwidth}
    \includegraphics[width=\columnwidth]{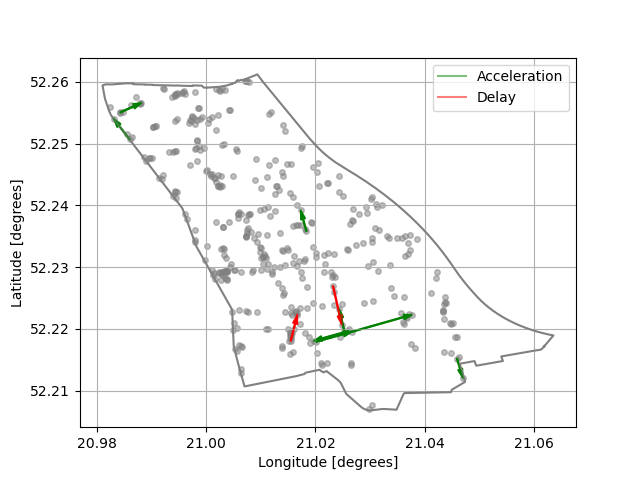}
    \caption{Delay 06:00-10:00}
    \label{fig:delay_morning}
\end{subfigure}
\begin{subfigure}[b]{0.33\textwidth}
    \includegraphics[width=\columnwidth]{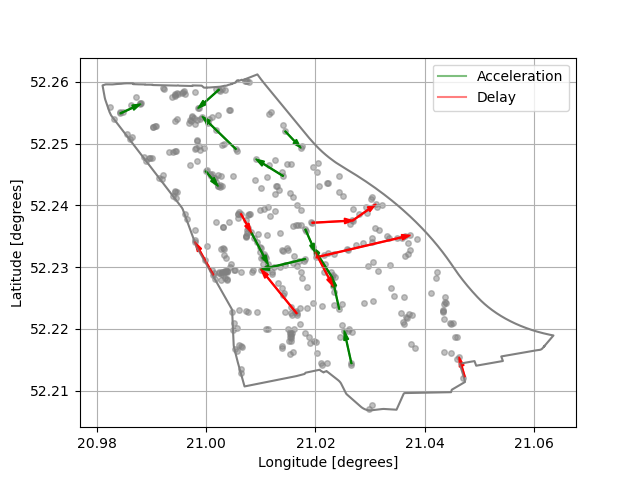}
    \caption{$\Delta$Delay 06:00-10:00}
    \label{fig:delta_morning}
\end{subfigure}
\begin{subfigure}[b]{0.33\textwidth}
    \includegraphics[width=\columnwidth]{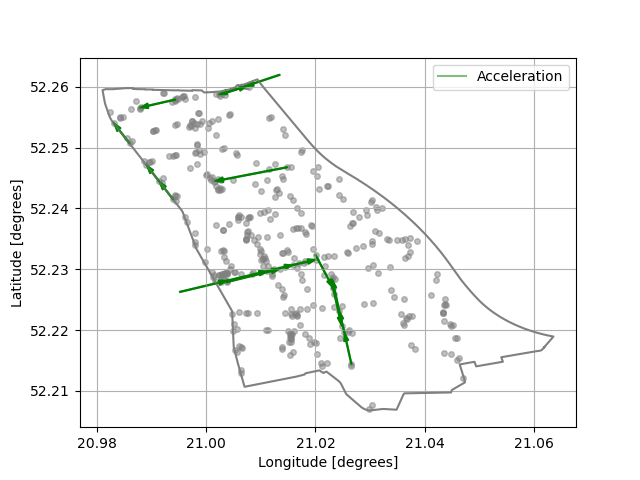}
    \caption{Delay 16:00-20:00}
    \label{fig:delay_evening}
\end{subfigure}
\begin{subfigure}[b]{0.33\textwidth}
    \includegraphics[width=\columnwidth]{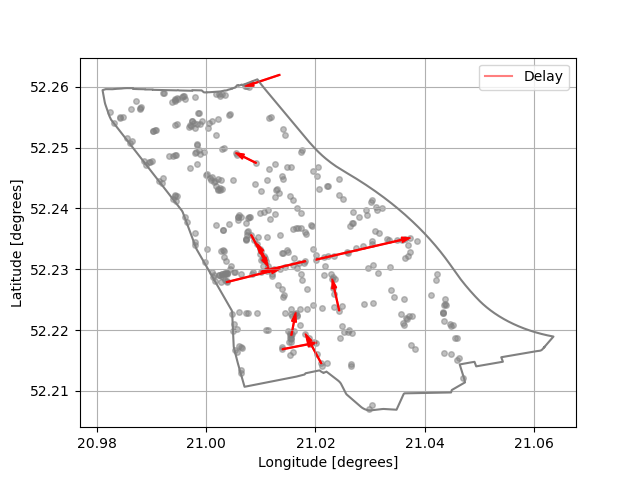}
    \caption{$\Delta$Delay 16:00-20:00}
    \label{fig:delta_evening}
\end{subfigure}
\caption{Changes in the stream of delay and $\Delta$delay values on 2\nth{1} December 2021. Edge-based approach.}
\label{fig:correlation_self}
\end{figure*}

To show how delay change detections can provide for more locally focused analysis, let us analyse detections observed during
two separate periods containing the morning and evening rush hours. %
Fig.~\ref{fig:correlation_self} compares detections 
between 6 am and 10 am (Fig.~\ref{fig:delay_morning} and Fig.~\ref{fig:delta_morning}), and  4 pm and 8 pm (Fig.~\ref{fig:delay_evening} and Fig.~\ref{fig:delta_evening}). 

Comparison of morning and evening delay detections (Fig.~\ref{fig:delay_morning} and Fig.~\ref{fig:delay_evening}) shows that some segments have acceleration or delay detected both in the morning and evening. That shows some segments of the traffic infrastructure with issues regardless of the time of the day.
Comparison of delay changes detections (Fig.~\ref{fig:delta_morning} and Fig.~\ref{fig:delta_evening}) shows that some segments changed the direction of detected delay as a direction of traffic jams changes between the morning and evening rush hours.
Finally, a comparison of both types of detectors shows segments with a delay detected by both types of detectors. In fact, there is a segment (in a rectangle defined by  21-21.02 Longitude and 52.21-52.22 Latitude),
 which illustrates the edges at which updates to static schedules could possibly be made.

\section{Conclusions and future works}
Delays in public transport may have a significant impact on mobility choices and discourage many citizens from the use of public transport services. However, delays reported based on vehicle location data may be caused both by inevitable temporal fluctuations and limited precision of GPS-based readouts. Furthermore, delays may occur due to short-term events such as a street temporarily partly blocked due to maintenance works.

To identify and focus on statistically significant delays, in this work rather than aggregating delays we propose the SDCD method detecting statistically significant changes in delays. 
The method makes it possible to detect both delays possibly arising or reduced in another part of PT system and propagated to the location of interest and delays arising or reduced at an edge of interest. Furthermore, we evaluate change detectors in terms of their usability to identify ADWIN as the most promising solution.

The method we propose is a part of the system integrating big data frameworks ensuring scalability of the solution and modelling environment including OpenTripPlanner instances. 
In the future, we will focus on how delay detections performed with the SDCD method can be aggregated to identify long term trends.

\subsubsection*{Acknowledgements}
This research has been supported by the CoMobility project. The CoMobility benefits from a 2.05 million\texteuro ~grant from Iceland, Liechtenstein and Norway through the EEA Grants. The aim of the project is to provide a package of tools and methods for the co-creation of sustainable mobility in urban spaces.

\vspace{-0.1cm}
\printbibliography
\end{document}